\pdfoutput=1

\documentclass[11pt]{article}

\usepackage{acl}

\usepackage{times}
\usepackage{latexsym}

\usepackage[T1]{fontenc}

\usepackage[utf8]{inputenc}

\usepackage{microtype}

\usepackage{inconsolata}

\usepackage{graphicx} 
\usepackage{wrapfig}
\usepackage{amsmath}
\usepackage{amssymb}
\usepackage{subcaption}
\usepackage{algorithm}
\usepackage{algorithmic}
\usepackage{booktabs}
\usepackage{enumitem}

\newcommand{\sizeof}[1]{\mathbf{#1}}

\title{Improving Reinforcement Learning from Human Feedback with \\Efficient Reward Model Ensemble}

\author{
    Shun Zhang$^{1}$, Zhenfang Chen$^{1}$, Sunli Chen$^{2}$, Yikang Shen$^{1}$, Zhiqing Sun$^{3}$, \and Chuang Gan$^{1,4}$ \\
    $^{1}$MIT-IBM Watson AI Lab,
    $^{2}$Tsinghua University,
    $^{3}$Carnegie Mellon University,
    $^{4}$UMass Amherst
}

\begin{document}
\maketitle

\begin{abstract}
Reinforcement Learning from Human Feedback (RLHF) is a widely adopted approach for aligning large language models with human values.
However, RLHF relies on a reward model that is trained with a limited amount of human preference data, which could lead to inaccurate predictions.
As a result, RLHF may produce outputs that are misaligned with human values.
To mitigate this issue, we contribute a reward ensemble method that allows the reward model to make more accurate predictions.
As using an ensemble of large language model-based reward models can be computationally and resource-expensive, we explore efficient ensemble methods including linear-layer ensemble and LoRA-based ensemble.
Empirically, we run Best-of-$n$ and Proximal Policy Optimization with our ensembled reward models, and verify that our ensemble methods help improve the alignment performance of RLHF outputs.
\end{abstract}

\section{Introduction}

Large language models (LLMs) \citep{vaswani_attention_2017} have become prominent in the field of artificial intelligence in solving a wide range of complex tasks in question answering \citep{ouyang_training_2022,guu_realm_2020}, code generation \citep{li_competition-level_2022,nijkamp_codegen_2023}, reasoning and planning \citep{kojima_large_2023,liu_llmp_2023}, and various other domains.
However, as large language models are trained using data from various sources, they may generate outputs that are biased, inappropriate, or even harmful, which are misaligned with human values.
Therefore, it is crucial to align large language models with human values for them to be safely deployed.

Recently, Reinforcement Learning from Human Feedback (RLHF) \citep{ouyang_training_2022} has shown to be a promising approach to mitigate the misalignment issue.
Concretely, RLHF first runs supervised fine-tuning (SFT) to train a large language model to generate responses that follow instructions.
It then trains a reward model using a preference dataset
that reflects human values.
Lastly, it runs reinforcement learning using the learned reward model to finetune the SFT model.
In this way, the finetuned model would generate sequences with higher rewards, which are presumably more aligned with human values.

The RLHF algorithm has demonstrated success in improving the alignment of large language models \citep{dubois_alpacafarm_2023,wu_fine-grained_2023}.
However, it also has some recognized issues.
As the reward model is trained on offline-collected preference data, its reward predictions may not be accurate on out-of-distribution data.
If we use reinforcement learning to optimize a large language model with an inaccurate reward model,
it may generate misaligned outputs with incorrectly high estimated rewards.
This problem is observed in the value alignment literature and is usually known as {\em reward hacking} or {\em reward overoptimization} \citep{amodei_concrete_2016,leike_ai_2017,gao_scaling_2022,casper_open_2023},
and is also a well-known problem in model-based offline reinforcement learning \citep{levine_offline_2020}.

In offline reinforcement learning, a prevalent strategy to mitigate reward overoptimization is to estimate rewards conservatively under uncertainty \citep{kumar_conservative_2020}.
Building upon this strategy, we consider an ensemble approach that employs a set of reward models to make better predictions,
in line with concurrent works \citep{coste_reward_2023,eisenstein_helping_2023}.
To achieve reward model ensemble, a straightforward approach is to train multiple reward models independently and then ensemble them.
However, this approach presents some challenges in our setting.
As reward models are usually based on large language models, training all the reward models and then loading all of them during inference time can be computationally expensive and resource-consuming.
Therefore, we contribute to designing efficient ensemble approaches.
Lastly, we empirically confirm the effectiveness of our ensemble methods using well-accepted evaluation benchmarks, AlpacaEval \citep{dubois_alpacafarm_2023} and MT-Bench \citep{zheng_judging_2023}.
The contributions of this paper are summarized as follows.
\begin{itemize}[noitemsep, topsep=0pt, parsep=0pt]
    \item We design reward ensemble algorithms that improve reward estimation accuracy, and hence improve the alignment of large language models. 
    \item We propose two ensemble approaches, 
    linear-layer ensemble and LoRA-based ensemble,
    that offer trade-offs between computational efficiency and alignment performance.
    \item We use the ensembled reward models for both Best-of-$n$ and Proximal Policy Optimization (PPO) algorithms, evaluate them on AlpacaEval and MT-Bench, and empirically confirm that RLHF with our ensembled reward models outperforms the standard RLHF algorithm.
\end{itemize}

\section{Background and Related Work}

\paragraph{Reinforcement learning from human feedback (RLHF).}
RLHF was originally considered in the TAMER framework \citep{knox_interactively_2009}, in which an agent learns the reward function from a human user's positive or negative feedback.
This setting is later considered in deep reinforcement learning \citep{christiano_deep_2017},
and recently employed for finetuning large language models \citep{ouyang_training_2022} to align the model's behavior with human values and preferences.
We overviewed the RLHF framework in the introduction and leave more details of the standard RLHF algorithm in Sec.~\ref{ap:rlhf}.

\paragraph{Ensemble models and uncertainty estimation.}
Model ensembling has been an accepted approach to improving a model's accuracy and estimating a model's uncertainty. 
\citet{lakshminarayanan_simple_2017} quantify predictive uncertainty in deep neural networks using ensembling, which performs better than Bayesian neural networks.
In reinforcement learning,
\citet{liang_reward_2022} use ensembled reward models to estimate the model's uncertainty for more informed exploration.
\citet{gleave_uncertainty_2022} use ensembled reward models for active learning.

Concurrently, \citet{coste_reward_2023} also use reward model ensemble to mitigate the reward model overoptimization problem and draw similar conclusions to our paper.
The key difference is that our work focuses on developing {\em efficient} ensemble approaches, since ensembling multiple independently-trained reward models can be expensive.
\citet{eisenstein_helping_2023} focus on comparing ensembling during pretraining and finetuning. 
\citet{zhai_uncertainty-penalized_2023} also consider LoRA-based ensemble,
while their work focuses on an uncertainty-penalized objective in RL-finetuning.
\citet{rame_warm_2024} consider a different approach of averaging the weights of multiple reward models instead of ensembling their predictions.
\citet{wang2023lora} use LoRA ensembles for improving predictive accuracy and uncertainty quantification.
\citet{ahmed2024scalable} propose a scalable ensemble approach that shares an encoder backbone but uses separate linear heads, achieving similar performance to full ensembling.

\paragraph{Offline reinforcement learning.}
Uncertainty estimation is also a key problem in offline reinforcement learning \citep{levine_offline_2020,janner_when_2019}.
For example, conservative Q-learning (CQL) \citep{kumar_conservative_2020} learns a conservative Q function to mitigate the overestimation bias for out-of-distribution state, action pairs.
Inspired by this algorithm, we also design an ensemble algorithm that uses conservative predictions by using the lower confidence bound of the ensembled predictions.

\section{Reward Model Ensemble}
\label{sec:method}

The reward model in RLHF is a (large language model-based) model that takes as input an instruction and a response, and outputs a reward prediction that indicates the alignment performance of the response.
In this section, we provide a formal descirption for our ensemble algorithms. 

\subsection{Architecture Design of Reward Model Ensemble} 
\label{sec:architecture}

\begin{figure*}[t]
    \centering
    \includegraphics[width=.85\textwidth]{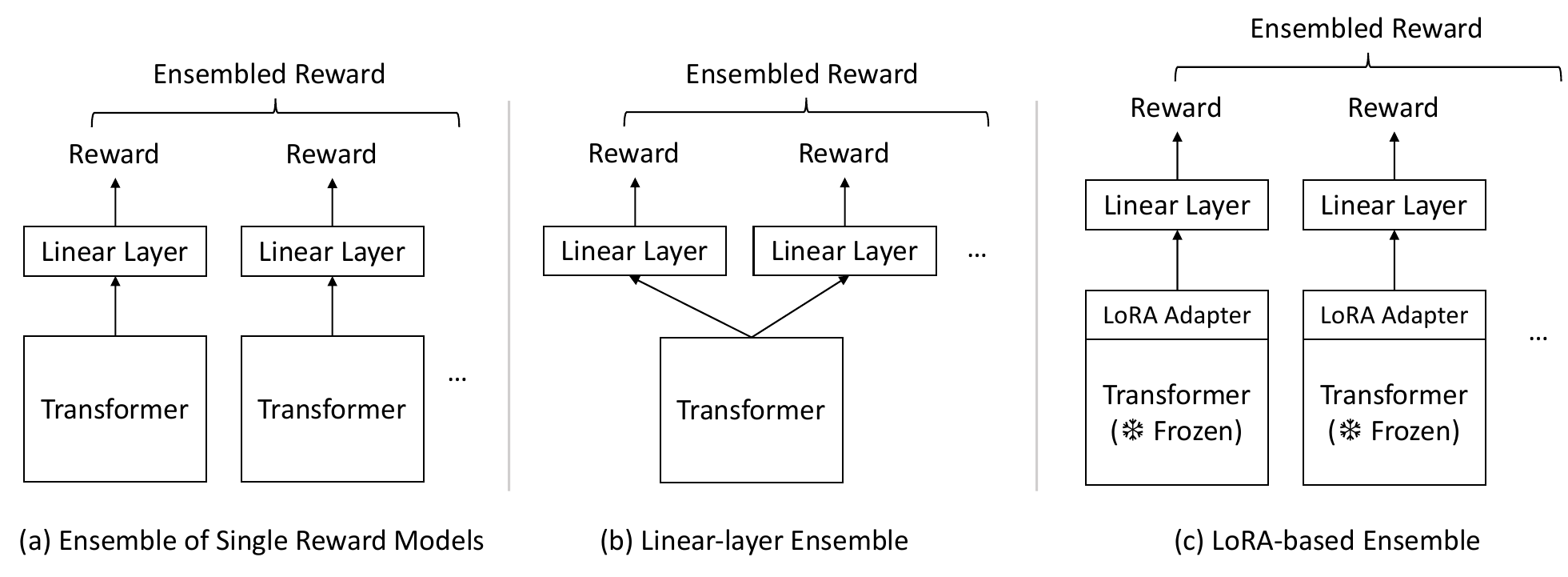}
    \vspace{-2pt}
    \caption{Illustration of the reward model ensemble algorithms. 
    }
    \label{fig:ensemble}
\end{figure*}

In the literature, reward models are commonly finetuned from pretrained large language models.
Following the convention \citep{dubois_alpacafarm_2023}, we assume a reward model is a Transformer model and a linear layer, 
where the linear layer takes as input the last hidden layer of the Transformer model, and outputs a reward value.

In this subsection, we discuss possible ways to ensemble multiple large language model-based reward models and discuss their advantages.
The ensemble algorithms are illustrated in Fig.~\ref{fig:ensemble} and the pseudocode is provided in Alg.~\ref{alg:ensemble}. 

\paragraph{Ensemble of single reward models.} A straightforward way to achieve reward model ensemble is to train multiple reward models independently using different random seeds.
At inference time, we simply load all the reward models and ensemble them.
However, such a method can be both expensive in training and inference --
during training, we need to run reward training for multiple times;
during inference, we need to load multiple large language model-based reward models simultaneously to GPUs, which can be resource-consuming.
Specifically, an ensemble of $k$ independently trained reward models needs to train and load $k(\sizeof{M} + \sizeof{L})$ parameters,
where $\sizeof{M}$ is the number of parameters of the Transformer model, and $\sizeof{L}$ is the number of parameters of a linear layer (shown in Fig.~\ref{fig:ensemble} (a)).

\paragraph{Linear-layer ensemble.}
To make the ensembling more efficient both in training and inference,
we can make all the ensembled models share the same Transformer model, while each ensembled model has its own linear layer that outputs its reward prediction.
During training, both the Transformer model and the linear layers of all the ensembled models are being trained.
Note that the Transformer model is the same for all the ensembled reward models.
In this way, a linear-layer ensemble model of size $k$ only requires $\sizeof{M} + k\sizeof{L}$ parameters (shown in Fig.~\ref{fig:ensemble} (b)).

\paragraph{LoRA-based ensemble.}
In linear-layer ensemble, allowing all the ensembled models to share the same Transformer model indeed reduces the total number of parameters requiring training.
However, it may considerably limit the diversity of the ensembled models.
Therefore, we allow each ensembled model to slightly finetune the Transformer model.
Specifically, each ensembled model trains its own linear layer, and a LoRA adapter \cite{hu2021lora} that is added to the Transformer model.
The LoRA adapter only has a small number of parameters and can be trained efficiently. 
(The background on LoRA is provided in Sec.~\ref{ap:lora}.)
In this way, a LoRA-based ensemble model of size $k$ requires $\sizeof{M} + k\sizeof{L} + k\sizeof{A}$ parameters, where 
$\sizeof{A}$ is the number of parameters of an adapter (shown in Fig.~\ref{fig:ensemble} (c)).

Empirically, we find that only LoRA-finetuning the Transformer model in the reward model does not perform well, as the Transformer model is not trained for reward prediction at all.
So in our experiments, we first finetune the Transformer model in the same way as linear-layer ensemble using a subset of preference data before ensemble model training (Line~\ref{line:lora_pretrain} in Alg.~\ref{alg:ensemble}).
We then use the rest of the data for ensemble training.
We provide more details on the dataset split in Sec.~\ref{sec:exp}.

\begin{algorithm}[t]
\captionsetup{font=small}
\caption{Reward Model Ensemble Algorithms}
\label{alg:ensemble}
\small{
\begin{algorithmic}[1]
\REQUIRE $k$: number of ensemble models, $M$: parameters of Transformer model
\STATE Initialize ensemble as empty set: $ensemble \gets \{\}$

\vspace{4pt}
\STATE $\blacktriangleright$ \textbf{Option 1: Ensemble of Single Reward Models}
\FOR{$i = 1$ \TO $k$}
    \STATE $M_i \gets \text{clone}(M)$
    \STATE Initialize linear layer with random parameters $L_i$
    \STATE $train(M_i \cup L_i)$
    \STATE Add model $M_i \cup L_i$ to $ensemble$
\ENDFOR

\vspace{4pt}
\STATE $\blacktriangleright$ \textbf{Option 2: Linear-layer Ensemble}
\STATE Initialize linear layer with random parameters $L_i$ for $i \in [0,k)$
\STATE Concurrently $train(M \cup L_i)$ for $i \in [0,k)$
\STATE Add $M \cup L_i, i \in [0,k)$ to $ensemble$

\vspace{4pt}
\STATE $\blacktriangleright$ \textbf{Option 3: LoRA-based Ensemble}
\STATE Finetune $M, L_i, \dots, L_k$ using linear-layer ensemble with a subset of data \label{line:lora_pretrain}
\FOR{$i = 1$ \TO $k$} \label{line:lora_start}
    \STATE Add LoRA adapter to the Transformer model $M$ with random parameters $A_i$
    \STATE $train(A_i \cup L_i)$
    \STATE Add $M \cup A_i \cup L_i$ to $ensemble$
\ENDFOR \label{line:lora_end}

\vspace{4pt}
\RETURN $ensemble$
\end{algorithmic}
}
\end{algorithm}

\subsection{Predictions of Ensembled Reward Models}
\label{sec:predict_alg}

Now we need to ensemble the predictions of different ensembled reward models.
We explore two algorithms for ensembling these predictions,
which use the mean predicted value and the lower confidence bound of the predicted values, respectively.

Let $R$ be the set of ensemble model predictions.
$R = \{r_1, r_2, \dots, r_k\}$.
{\bf Mean value prediction} simply uses $\text{mean}(R)$, the mean value of the ensemble reward model prediction.
This is inherently a lower-variance estimation of the reward. 
On the other hand, {\bf lower confidence bound (LCB)} is a conservative estimation of the reward. It considers the standard deviation of the ensemble model predictions,
defined as 
$\text{LCB}(R) = \text{mean}(R) - \beta \cdot \text{std}(R)$,
where $\beta$ is a hyperparameter.
However, we empirically find that the performance of LCB is comparable to that of mean value prediction.

\section{Empirical Evaluation}
\label{sec:exp}

\begin{figure*}[t]
\centering

\begin{subfigure}{.32\textwidth}
\centering
\includegraphics[width=\textwidth]{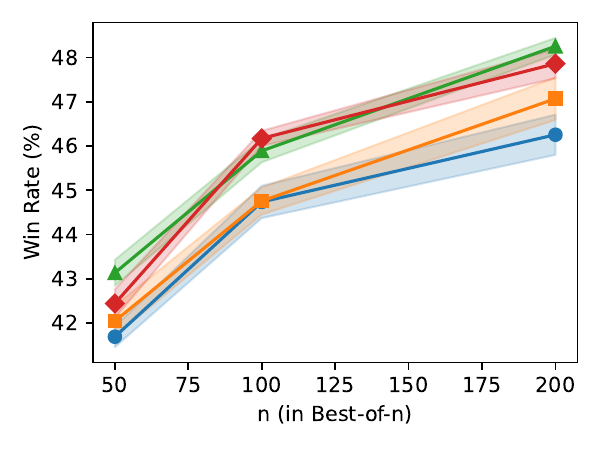}
\caption{Best-of-$n$ results.}
\label{fig:best_of_n}
\end{subfigure}
\begin{subfigure}{.32\textwidth}
\centering
\includegraphics[width=\textwidth]{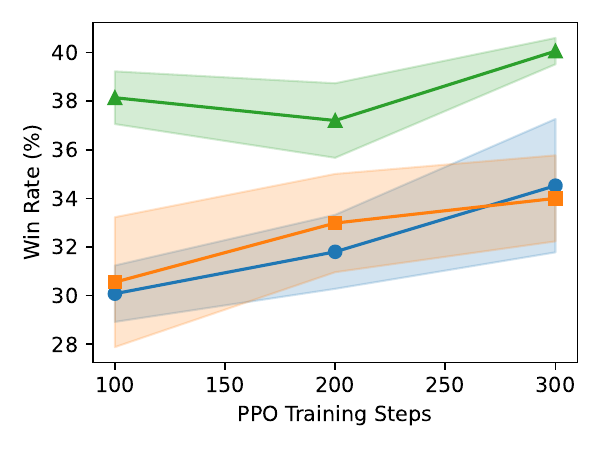}
\caption{PPO results.}
\label{fig:ppo}
\end{subfigure}
\begin{subfigure}{.33\textwidth}
\centering
\includegraphics[clip, trim=0 -30mm 0 0, width=\textwidth]{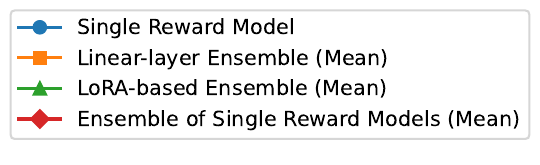}
\end{subfigure}

\caption{Win rates of model responses using Best-of-$n$ and PPO on AlpacaEval. Different lines represent different reward ensemble algorithms. The shaded areas represent standard errors.}
\label{fig:main}
\end{figure*}

In this section, we empirically answer the following questions.
{\bf Q1:} Compared with using a single reward model, do ensembled reward models help improve alignment performance in RLHF? 
{\bf Q2:} Which ensemble architecture in Sec.~\ref{sec:architecture} has the best performance?
{\bf Q3:} Which prediction algorithm in Sec.~\ref{sec:predict_alg} has a better performance?

\paragraph{Algorithms.} We consider using the ensembled reward models with Best-of-$n$ and Proximal Policy Optimization (PPO) \citep{schulman_proximal_2017}, which are standard approaches in RLHF \citep{ouyang_training_2022,dubois_alpacafarm_2023,coste_reward_2023}.
Specifically,
\textbf{Best-of-$n$} generates $n$ samples from the SFT model for each input and selects the sample with the highest predicted reward.
\textbf{Proximal Policy Optimization (PPO)} is a reinforcement learning algorithm that finetunes the SFT model using the reward model. 

For each reward ensemble method, we conduct the experiments multiple times using different random seeds. Specifically, we repeat the experiments 10 times for Best-of-$n$ and 5 times for PPO.
For all experiments except experiments without reward ensemble, we use three reward models for ensembling ($k=3$).

\paragraph{Models.} We use the pretrained models provided in AlpacaFarm \citep{dubois_alpacafarm_2023}.
Specifically, we use \texttt{SFT10k} as the base model for generation, which is a Llama-7b model \citep{touvron2023llama} finetuned on the \texttt{alpaca\_instructions} dataset.
So the model can follow instructions, while it has not been aligned with human preferences.
To be consistent with \citet{dubois_alpacafarm_2023}, the Transformer model in our reward model is also initialized using \texttt{SFT10k}, which is not yet trained for reward prediction.

\paragraph{Datasets.} We use the AlpacaFarm datasets \citep{dubois_alpacafarm_2023} for training and evaluation, which provide utilities to evaluate the alignment performance of the model outputs using GPT-4 APIs and can easily compare our models with other benchmarking models and algorithms.
We train all the reward models using both \texttt{alpaca\_noisy\_multi\_preference} and \texttt{alpaca\_human\_preference} datasets,
and use \texttt{alpaca\_farm\_evaluation} for evaluation.

Specifically, for LoRA-based ensemble, we use the two training datasets for different phases:
We first use \texttt{alpaca\_noisy\_multi\_preference} to fully finetune the Transformer model with $k$ linear layers, in the same way as linear-layer ensemble (Line~\ref{line:lora_pretrain} in Alg.~\ref{alg:ensemble}), as we find fully-finetuning the Transformer is necessary for it to make reward predictions. Then we use \texttt{alpaca\_human\_preference} for training the linear layers and the adapters for the ensemble members (Line~\ref{line:lora_start}-\ref{line:lora_end} in Alg.~\ref{alg:ensemble}).

\paragraph{Results.}
Our main results are reported in Figure~\ref{fig:main}. 
For Best-of-$n$, we choose $n=50, 100, 200$.
For PPO, we evaluate checkpoints at every 100 training steps. %
To evaluate the alignment performance, we use the win rate metric provided in AlpacaEval (shown as the vertical axis), which measures the chances that our methods' outputs are preferred by GPT-4 compared with a GPT-3 generated baseline. 
All the ensemble approaches use the mean value prediction, which uses the mean of the predicted rewards of the ensemble members as their final predictions. 

Overall, we find that the win rates are consistently higher when using reward method ensemble (answering {\bf Q1}). 
For Best-of-$n$, both ensemble of single reward models and LoRA-based ensemble have the best performance. 
For PPO, we are unable to run ensemble of single reward models as it requires loading multiple reward models during PPO training.
Nonetheless, we find LoRA-based ensemble has the best performance.

We also conduct the experiments on MT-Bench \citep{zheng_judging_2023}, which is a benchmark for multiturn questions.
We evaluate our PPO-trained models with the most training steps,
and report the alignment scores after the first turn, after the second turn, and the average of both.
The results are reported in Table~\ref{tab:mt_bench}, and our findings are consistent with the AlpacaEval results.
The results on both benchmarks suggest that, although LoRA does not fully finetune the Transformer models, it is effective for reward model ensemble and can improve the alignment performance (answering {\bf Q2}).

In terms of the prediction methods, we find that the mean reward prediction and LCB have similar performance. Detailed results are presented in Sec.~\ref{ap:more_results} (answering {\bf Q3}).

\begin{table}[t]
\centering
\small{
\begin{tabular}{lrrr}
\toprule
\text{Ens.\ Method} & \text{First Turn} & \text{Second Turn} & \text{Average} \\
\midrule
\text{Ens.\ of Single} & 4.70 $\pm$ 0.12 & 3.63 $\pm$ 0.22 & 4.16 $\pm$ 0.14 \\
\text{Linear-layer} & 4.73 $\pm$ 0.22 & 3.67 $\pm$ 0.19 & 4.20 $\pm$ 0.10 \\
\text{LoRA-based} & 4.86 $\pm$ 0.09 & 3.84 $\pm$ 0.21 & 4.35 $\pm$ 0.12 \\
\bottomrule
\end{tabular}
}
\caption{Alignment scores on MT-Bench for different ensemble methods.}
\label{tab:mt_bench}
\end{table}

\section{Discussion and Conclusion}

In summary, our paper presents a novel approach to enhancing the alignment of large language models through efficient reward model ensemble in RLHF.
Specifically, the LoRA-based ensemble method demonstrates effectiveness under computational constraints.
In future work, we will extend this approach to other steps of LLM training and inference, such as sample-efficient training of reward models \cite{gleave_uncertainty_2022}.

\bibliography{zotero,extra_references}

\appendix

\section{Preliminaries}

For the completeness of the paper, we provide more background details on reinforcement learning from human feedback and LoRA finetuning in this section.

\subsection{Reinforcement Learning from Human Feedback}
\label{ap:rlhf}

Reinforcement learning from human feedback (RLHF) follows a three-step process: supervised fine-tuning (SFT), reward modeling, and reinforcement learning using the learned reward model.

In this paper, we focus on the second step of RLHF, which trains a reward model that reflects human preferences.
We denote the reward model by $r_\theta$, parameterized by $\theta$.
To train the reward model, we have a preference dataset, $\mathcal{D} = \{(x,y_{w},y_{l}), \dots\}$, where $x$ is a context (a question or an instruction); $y_{w}$ is a preferred output, and $y_{l}$ is a less preferred output.
The reward model is then trained to predict the preference score $r_{\theta}(x,y)$ for an input-output pair, where a larger score indicates that the output is more preferred by a human.
The loss function for reward model training is
\begin{align}
\text{loss} (\theta) = &-\mathbb{E}_{(x,y_{w},y_{l})\sim\mathcal{D}} \nonumber \\
&\left[ \log \left( \sigma \left( r_{\theta} (x, y_w) - r_{\theta} (x, y_l) \right) \right) \right],
\end{align}
where $\sigma$ represents the sigmoid function. This approach effectively captures humans' preferences, so that $r_\theta$ predicts higher rewards for responses that are preferred by humans.

In the last step of RLHF, we can finetune the supervised finetuned (SFT) model using reinforcement learning, typically using the Proximal Policy Optimization (PPO) algorithm.
In essence, PPO iteratively improves the policy by simultaneously minimizing the divergence between new and old policies and maximizing the expected cumulative rewards. We refer readers to find algorithm details in the original paper \citep{schulman_proximal_2017}.

\subsection{LoRA Finetuning}
\label{ap:lora}

When finetuning a large language model, finetuning all the parameters can be computationally expensive and resource-demanding.
To this end, Low-Rank Adaptation (LoRA) \citep{hu2021lora} is a well-accepted algorithm to efficiently finetune a pretrained large language model.
Concretely, for each Transformer layer, LoRA learns
\begin{equation}
\Delta W = A_1 A_2,
\end{equation}
where $\Delta W$ is the change applied to the weight matrix $W$ of a transformer layer, and $A_1$ and $A_2$ are smaller matrices.
Let the dimension of $W$ be $d_1 \times d_2$.
Then the dimension of $A_1$ is $d_1 \times r$ and the dimension of $A_2$ is $r \times d_2$, where $r$ is the rank of the decomposition, which is smaller than $d_1$ and $d_2$.

When applying LoRA to a transformer layer, the modified weight matrix $W'$ is used in the forward pass as follows,
\begin{equation}
W' = W + \Delta W,
\end{equation}
where $W$ is the original weight matrix of the transformer layer.
It is worth noting that during training, only the matrices $A_1$ and $A_2$ are updated. The original weights $W$ remain frozen.
This approach requires training only a substantially smaller set of parameters in matrices $A_1$ and $A_2$, compared to the original weights $W$, while all the model weights will be influenced during training.

\section{Additional Empirical Results}
\label{ap:more_results}

\begin{figure}[h]
\centering

\begin{subfigure}{.35\textwidth}
    \centering 
    \vspace{5pt}
    \includegraphics[width=\textwidth]{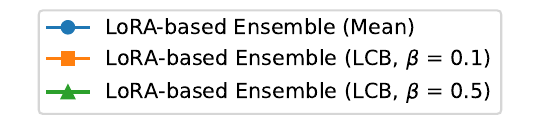}
\end{subfigure}
\begin{subfigure}{.4\textwidth}
    \centering
    \includegraphics[width=0.85\textwidth]{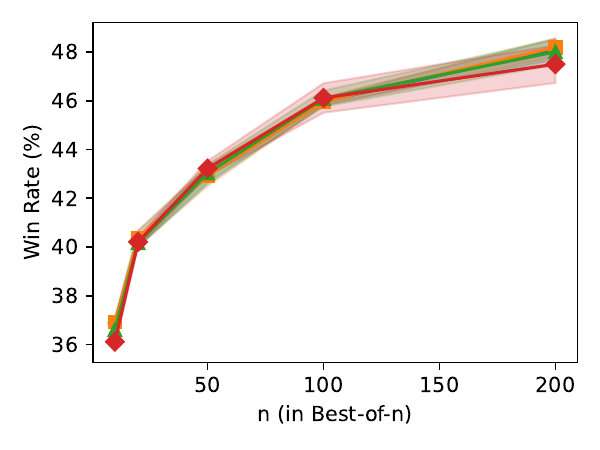}
    \label{fig:bon_lora}
\end{subfigure}

\caption{Lora-based ensemble with LCB and different $\beta$ values.}
\label{fig:lora_multi_betas}

\begin{subfigure}{.35\textwidth}
    \centering 
    \vspace{5pt}
    \includegraphics[width=\textwidth]{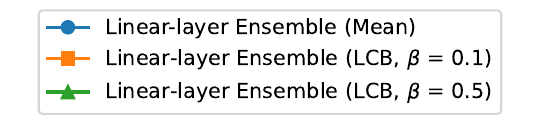}
\end{subfigure}
\begin{subfigure}{.4\textwidth}
    \centering
    \includegraphics[width=0.85\textwidth]{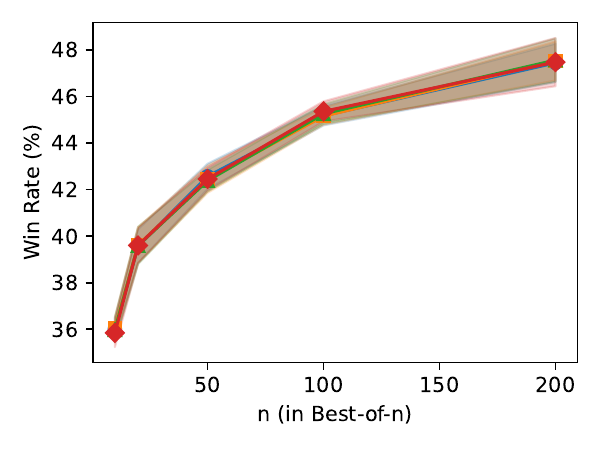}
    \label{fig:bon_linear}
\end{subfigure}
\caption{Linear-layer ensemble with LCB and different $\beta$ values.}
\label{fig:linear_multi_betas}
\end{figure}

In addition to our results in the main paper, we also explored different prediction algorithms,
including using the mean and lower confidence bound (LCB) of the reward predictions of the ensembled models.
We perform the experiments on AlpacaEval.
The differences between these prediction algorithms are insignificant, as shown in Figures~\ref{fig:lora_multi_betas} and \ref{fig:linear_multi_betas}.

\section{Hyperparameters}

We include the key hyperparameters for our experiments in the following tables.

\begin{table}[!htbp]
\centering
\small{
\begin{tabular}{ll}
\toprule
\textbf{Hyperparameter}                & \textbf{Value}   \\
\midrule
seed                                   & 42, 43, $\dots$, 51               \\
num\_train\_epochs                     & 1                \\
gradient\_accumulation\_steps          & 2                \\
learning\_rate                         & 3e-6             \\
weight\_decay                          & 0.0              \\
warmup\_ratio                          & 0.03             \\
optimizer                              & adamw\_torch                    \\
lr\_scheduler\_type                    & cosine           \\
\bottomrule
\end{tabular}
}
\caption{Parameters for reward modeling.}
\end{table}

\begin{table}[!htbp]
\centering
\small{
\begin{tabular}{ll}
\toprule
\textbf{Hyperparameter}                & \textbf{Value}   \\
\midrule
seed                                   & 42, 43, $\dots$, 51              \\
num\_train\_epochs                     & 1                \\
gradient\_accumulation\_steps          & 2                \\
learning\_rate                         & 5e-5             \\
weight\_decay                          & 0.0              \\
warmup\_ratio                          & 0.03             \\
optimizer                              & adamw\_torch                    \\
lr\_scheduler\_type                    & constant         \\
\bottomrule
\end{tabular}
}
\caption{Parameters for reward modeling with LoRA.}
\end{table}

\begin{table}[!htbp]
\centering
\small{
\begin{tabular}{ll}
\toprule
\textbf{Hyperparameter}                & \textbf{Value}                  \\
\midrule
rollout\_batch\_size                   & 64                              \\
step\_batch\_size                      & 32                              \\
learning\_rate                         & 1e-5                            \\
warmup\_steps                          & 5                               \\
epoch\_num                             & 2                               \\
optimizer                              & adamw\_torch                    \\
kl\_divergence\_coefficient              & 0.02                            \\
value\_function\_coefficient             & 0.1                             \\
\bottomrule
\end{tabular}
}
\caption{PPO parameters.}
\end{table}

\begin{table}[!htbp]
\centering
\small{
\begin{tabular}{ll}
\toprule
\textbf{Hyperparameter}         & \textbf{Value} \\
\midrule
temperature                & 1.0            \\
max\_new\_tokens           & 300            \\
top\_p                     & 0.9            \\
\bottomrule
\end{tabular}
}
\caption{Decoding parameters for Best-of-$n$.}
\end{table}

\begin{table}[!htbp]
\centering
\small{
\begin{tabular}{ll}
\toprule
\textbf{Hyperparameter}         & \textbf{Value} \\
\midrule
temperature                & 0.7            \\
max\_new\_tokens           & 300            \\
top\_p                     & 0.9            \\
\bottomrule
\end{tabular}
}
\caption{Decoding parameters for PPO.}
\end{table}

\end{document}